\documentclass{article}
\usepackage[eandd,preprint]{neurips_2026}
\usepackage[utf8]{inputenc}
\usepackage[T1]{fontenc}
\usepackage{hyperref}
\usepackage{url}
\usepackage{booktabs}
\usepackage{amsfonts}
\usepackage{nicefrac}
\usepackage{microtype}
\usepackage{xcolor}
\usepackage{graphicx}
\usepackage{multirow}
\usepackage{array}
\usepackage{enumitem}
\usepackage{amsmath}
\usepackage{listings}
\usepackage{tcolorbox}
\usepackage{tabularx}
\tcbuselibrary{breakable,skins}
\newcommand{\ir}{\textbf{ImmigrationReason}}

\title{\ir{}: A Structured Dataset of U.S.\ Immigration Appeals for Legal Reasoning Research}

\author{%
  Amirhossein Afsharrad \\
  Stanford University \\
  \texttt{afsharrad@stanford.edu} \\
  \And
  Seyed Shahabeddin Mousavi \\
  Stanford University \\
  \texttt{ssmousav@cs.stanford.edu}
}

\begin{document}

\maketitle

\begin{abstract}
Most legal NLP resources draw from federal case law and focus on coarse
classification, leaving administrative adjudication, where the vast
majority of government decisions occur, essentially unaddressed.
We introduce \ir{}, a large-scale structured dataset derived from 12,375
non-precedent decisions of the U.S.\ Citizenship and Immigration Services
(USCIS) Administrative Appeals Office (AAO) spanning 2005 to 2026.
Each record captures the applicable legal framework, per-criterion
evidence-sufficiency findings under a five-category label, verbatim
adjudicator-criticism quotes, all citations, and final dispositions,
alongside high-quality Claude-transcribed source text.
Extraction quality is validated through a three-pass pipeline combining
two independent modalities with comparison-prompt adjudication by
Opus~4.7, and verified by domain experts on a 500-record sample.
The dataset documents nearly 9,000 verbatim instances of AAO-identified
legal errors, spans a natural legal-regime transition (the 2016
\emph{Dhanasar} rule change), and covers 21 years of adjudication.
We analyze the dataset in detail and outline research directions it
enables, from outcome prediction and adjudicator-error analysis to
agent design for high-stakes regulatory domains.
\end{abstract}

%% ====================================================================
\section{Introduction}
\label{sec:intro}
%% ====================================================================

Legal reasoning is emerging as one of the most consequential and
challenging frontiers for large language models (LLMs).
Commercial deployments now target legal research, contract drafting,
compliance analysis, and advisory tasks~\citep{magesh2024hallucination,dahl2024large}.
Yet evaluating and improving legal-reasoning capabilities requires data
resources that reflect the actual complexity of legal work: multi-step
evidence weighing against evolving doctrinal standards, citation-grounded
argumentation, and the detection of subtle errors in a counterpart's reasoning.
Existing legal NLP datasets~\citep{guha2023legalbench,chalkidis2022lexglue,fei2023lawbench}
have made important contributions, but they are predominantly built from
federal case law and structured as coarse-grained classification tasks.
The deeper layer of legal decision-making, namely administrative
adjudication, where federal agencies determine individual rights and
entitlements under regulatory frameworks, is almost entirely absent.

Administrative agencies make orders of magnitude more decisions annually
than the federal courts, and they do so under doctrinal standards that are
categorically different from common-law case law.
Among these, U.S.\ employment-based immigration adjudication is
particularly significant: USCIS processes millions of petitions per year;
the Administrative Appeals Office (AAO) issues public,
richly reasoned decisions on denied petitions; and USCIS has explicitly
acknowledged deploying AI tools in its adjudication
process~\citep{dhs2026aicatalog}, making independent evaluation both
urgent and practically important.

We introduce \ir{}, a large-scale structured dataset built from 12,375
non-precedent AAO decisions spanning 2005 to 2026.
Unlike raw legal corpora, \ir{} provides deep structured annotations at
three levels: decision-level metadata (posture, frameworks, petitioner
field), per-legal-issue analysis (issue type, reasoning, conclusion), and
per-criterion findings under a five-category label (separately tracking
both the officer's and the AAO's determination for each prong).
The dataset is distinguished by three features that do not exist in any
prior legal NLP resource:
(1)~per-criterion evidence-sufficiency labels derived from a senior
administrative tribunal,
(2)~explicit labels of adjudicator errors (verbatim AAO criticisms of
originating officers), and
(3)~a built-in temporal legal-regime transition (the December 2016
\emph{Dhanasar} rule change) that provides natural out-of-distribution
test conditions.

%% ====================================================================
\section{Related Work}
\label{sec:related}
%% ====================================================================

\paragraph{Legal NLP datasets and benchmarks.}
LegalBench~\citep{guha2023legalbench} is the most comprehensive legal
reasoning benchmark, covering 162 tasks, but is U.S.\ case-law-centric
and classification/span-extraction oriented.
LexGLUE~\citep{chalkidis2022lexglue} provides seven legal classification
datasets across jurisdictions.
LawBench~\citep{fei2023lawbench} and LEXam~\citep{louis2024lexam} focus
on Chinese and Swiss law.
Pile of Law~\citep{hendersonkrass2022pile} assembles 256~GB of raw legal
text as a pretraining corpus.
CaseHOLD~\citep{zheng2021does} provides citation-prediction labels from
U.S.\ federal case opinions.
None of these provide structured per-criterion findings, adjudicator-error
annotations, or administrative-law coverage.

\paragraph{Immigration law specifically.}
\citet{barale2023refugee} provide a small Canadian refugee-law retrieval
dataset (400 cases).
\citet{chen2017predicting} study asylum-outcome prediction as a pre-LLM
classification task.
Neither covers U.S.\ employment-based adjudication or provides structured
annotations.

\paragraph{Document digitization and OCR.}
Pile of Law~\citep{hendersonkrass2022pile} and prior legal NLP work
have largely relied on classical OCR pipelines (e.g., Tesseract,
PyMuPDF) for scanned legal documents.
As a baseline comparison for our corpus, we measured the quality of
classical PyMuPDF extraction against our Claude-powered transcription
across 12,375 paired pre- and post-2017 decisions.
The results, summarized in Table~\ref{tab:ocr}, confirm the expected
finding: LLM-based transcription substantially outperforms classical OCR
on noisy scanned documents, recovering 3.3$\times$ more footnotes,
eliminating 99.9\% of noise tokens, and achieving up to 5$\times$ more
content in severely truncated pre-2017 cases.
This gap motivates our choice to release Claude-powered transcriptions
rather than raw OCR as the canonical source text.

\paragraph{Legal AI deployment and hallucination.}
\citet{magesh2024hallucination} and \citet{dahl2024large} document
significant citation-hallucination rates in commercial legal AI, and
\citet{place2025jurisdiction} show jurisdiction-specific variation.
\ir{} provides a substrate for studying these phenomena in a context
where the AI is already being deployed and enables the evaluation of legal argument cogency in a real-world setting.

%% ====================================================================
\section{Background: AAO Adjudication}
\label{sec:background}
%% ====================================================================

The USCIS Administrative Appeals Office reviews Form~I-140
employment-based immigrant visa petitions after denial by a USCIS
Service Center.
The AAO conducts \emph{de novo} review~\citep{christo2015} and issues
non-precedent decisions publicly.

\textbf{EB-1A (Extraordinary Ability).}
Governed by 8~C.F.R.~\S~204.5(h)(3), which provides ten evidentiary
criteria; a petitioner must meet at least three.
Since \emph{Kazarian v.\ USCIS}~\citep{kazarian2010}, AAO applies a
two-step analysis: count qualifying criteria, then make a final-merits
determination of sustained national or international acclaim.

\textbf{EB2-NIW (National Interest Waiver).}
Since December 2016, governed by the three-prong
\emph{Dhanasar}~\citep{dhanasar2016} framework: (1)~substantial merit
and national importance; (2)~well-positioned to advance the endeavor;
(3)~on balance, beneficial to waive the job-offer requirement.
Before December 2016, the NYSDOT framework~\citep{nysdot1998} applied,
requiring different showings.
This transition is a central feature of \ir{}: it creates a hard temporal
boundary in the data's legal structure.

The decision body follows a consistent structure: introductory summary
(including the outcome announcement), LAW section (statutory and
regulatory framework), ANALYSIS section (per-criterion discussion), and
ORDER block.

%% ====================================================================
\section{Dataset Construction}
\label{sec:construction}
%% ====================================================================

\subsection{Source Collection and Filtering}

We scraped all publicly available AAO non-precedent decisions in the
EB-1A (form code B2203) and NIW (form code B5203) categories from the
USCIS website, yielding 13,520 PDFs spanning January 2005 to March 2026.
After filtering out O-1 filings (small and structurally different),
documents under 2,000~characters after text extraction, and a small
number of duplicates, the final corpus contains \textbf{12,375 decisions}.

\subsection{Source Text Preparation}
\label{sec:text}

PDFs fall into two quality categories. Post-2017 documents are
text-native (USCIS began producing them with embedded text); PyMuPDF
extracts these near-losslessly.
Pre-2017 documents are predominantly scanned images; classic OCR
produces noisy results with garbled citations, missing footnotes, and
truncated content.
For all documents, we produce a Claude-transcribed Markdown
version using Claude Sonnet~4.6 with a detailed transcription
prompt (see Appendix~\ref{app:prompt_transcription}).
Table~\ref{tab:ocr} quantifies the improvement.

\begin{table}[t]
  \caption{OCR quality comparison across 12,375 paired records.
    Claude-powered transcription dramatically reduces noise and recovers
    substantially more structural content, especially in pre-2017
    scanned documents.}
  \label{tab:ocr}
  \centering\small
  \begin{tabular}{lrrl}
    \toprule
    \textbf{Feature} & \textbf{Legacy OCR} & \textbf{Vision OCR}
      & \textbf{Change} \\
    \midrule
    OCR noise tokens (e.g.\ \texttt{tj} for \S) & 8,369 & 7 & $-$99.9\% \\
    Footnote markers preserved & 20,086 & 66,690 & $+$232\% \\
    \texttt{[REDACTED]} markers & 13,040 & 206,553 & $+$1,485\% \\
    CFR citations & 75,253 & 112,587 & $+$50\% \\
    I\&N Dec.\ citations & 47,465 & 60,745 & $+$28\% \\
    Max doc recovery (single doc) & 9,292 chars & 66,938 chars & $+$620\% \\
    \midrule
    Docs $\ge$85\% of legacy length & -- & 12,078 & 97.6\% \\
    \bottomrule
  \end{tabular}
\end{table}

\begin{figure}[t]
  \centering\includegraphics[width=\linewidth]{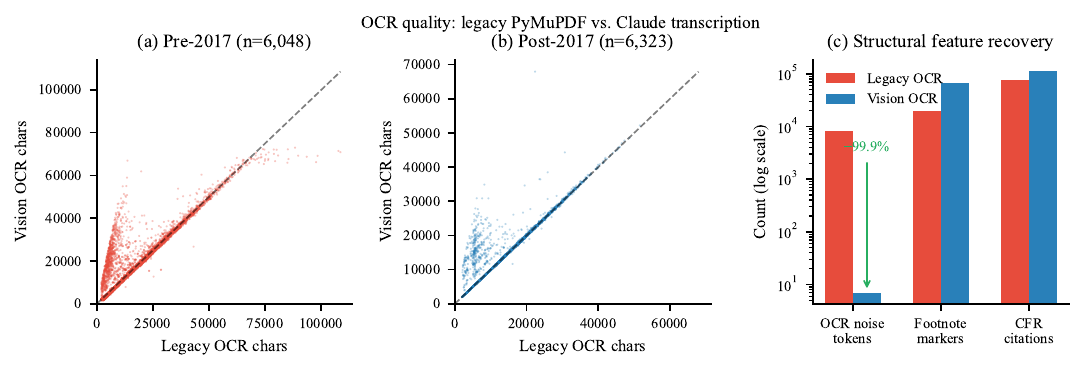}
  \caption{OCR quality before and after Claude-powered transcription.
    (a,b) Pre-2017 documents show the most improvement; many had been
    severely truncated by classic OCR.
    (c) Corpus-wide structural-feature recovery across key signal types.}
  \label{fig:ocr}
\end{figure}

\subsection{Extraction Schema}
\label{sec:schema}

Each decision is decomposed into a typed \texttt{Decision} record with
three nested levels.
The full Pydantic schema is released with the dataset; we summarize the
key fields here.

\textbf{Decision level} (one per document):
\texttt{filename\_stem}, \texttt{decision\_date}, \texttt{posture}
(8~types: appeal, motion-to-reopen, motion-to-reconsider, combined
motion, appeal-of-approval, certification, remand-review, other),
\texttt{visa\_category} (5~types), \texttt{originating\_office},
\texttt{petitioner\_field} (free text), \texttt{petitioner\_type},
\texttt{aao\_overall\_disagreement\_with\_director} (4~types),
\texttt{aao\_specific\_criticisms\_of\_director} (list of verbatim
quotes), \texttt{citations} (typed list), \texttt{final\_orders},
and \texttt{extraction\_confidence}.

\textbf{LegalIssue level} (list per decision, median 2, max 7):
\texttt{issue\_type} (17~values covering substantive frameworks and
cross-cutting employment-based issues), \texttt{framework\_citation},
\texttt{aao\_conclusion} (5~types), \texttt{aao\_reasoning\_summary},
\texttt{aao\_reasoning\_quotes} (verbatim list).

\textbf{Finding level} (list per issue, median 3, max 14):
\texttt{prong\_id}, \texttt{prong\_name},
\texttt{petitioner\_evidence\_types} (list),
\texttt{petitioner\_arguments\_summary},
\texttt{director\_finding} (4~states),
\texttt{aao\_finding} (5~states; see below),
\texttt{aao\_reasoning\_summary}, \texttt{aao\_reasoning\_quotes},
\texttt{aao\_agrees\_with\_director}, and
\texttt{aao\_specific\_criticism\_of\_director} (verbatim).

\paragraph{The five-category finding label.}
The \texttt{aao\_finding} field is the most novel structural element.
It takes one of five values:
\texttt{met} (AAO found the criterion satisfied),
\texttt{not\_met} (AAO found it unsatisfied),
\texttt{reserved} (AAO explicitly declined to reach the issue because
another issue was dispositive, per \emph{INS v.\ Bagamasbad}~\citep{bagamasbad1976}),
\texttt{waived\_by\_petitioner} (petitioner did not raise the issue on
appeal and it was deemed abandoned), and
\texttt{not\_addressed} (neither party substantively discussed the issue).
As Table~\ref{tab:corpus} shows, 19.8\% of all findings are
\texttt{reserved} and 3.4\% are \texttt{waived}; a binary \{met, not\_met\}
scheme would miscategorize 23\% of all finding instances, erasing signals
critical to understanding adjudication logic and producing incorrect
ground-truth labels for any downstream task.

\begin{table}[t]
  \caption{ImmigrationReason dataset statistics.}
  \label{tab:corpus}
  \centering\small
  \begin{tabular}{lrl}
    \toprule
    \textbf{Attribute} & \textbf{Value} & \textbf{Notes} \\
    \midrule
    Total decisions & 12,375 & 2005--2026 \\
    NIW (Dhanasar framework) & 3,484 & post-2016 \\
    NIW (NYSDOT framework) & 1,280 & pre-2016 \\
    EB-1A (Kazarian) & 6,486 & step-1 + step-2 issues \\
    EB-2 threshold only & 2,626 & no NIW prong analysis \\
    Other / procedural & 499 & \\
    \midrule
    Sustained & 674 & 5.4\% \\
    Remanded & 975 & 7.9\% \\
    Dismissed/rejected & 10,373 & 83.8\% \\
    Other & 353 & \\
    \midrule
    Pre-2017 (scanned PDFs) & 6,051 & \\
    Post-2017 (text-native PDFs) & 6,324 & \\
    \midrule
    Total per-criterion findings & 45,290 & \\
    \quad \textit{met} & 8,139 & 18.0\% \\
    \quad \textit{not\_met} & 25,454 & 56.2\% \\
    \quad \textit{reserved} & 8,972 & 19.8\% \\
    \quad \textit{waived\_by\_petitioner} & 1,527 & 3.4\% \\
    \quad \textit{not\_addressed} & 1,198 & 2.6\% \\
    \midrule
    AAO fully agreed with director & 5,678 & 45.9\% \\
    AAO partially disagreed & 3,650 & 29.5\% \\
    AAO fully disagreed (reversed) & 1,487 & 12.0\% \\
    Procedural / N/A & 1,560 & 12.6\% \\
    \midrule
    Verbatim AAO criticism quotes & $\sim$9,000 & across 4,377 records \\
    Unique petitioner fields (free text) & $\sim$1,000 & \\
    Unique issue types & 17 & \\
    Median citations per decision & 10 & \\
    \bottomrule
  \end{tabular}
\end{table}

\subsection{Extraction Pipeline}
\label{sec:pipeline}

We extract structured records using Claude Sonnet~4.6 via forced tool
use~\citep{anthropic2024tooluse} with Pydantic~\citep{pydantic2023}
validation and retry-on-validation-feedback (up to 3 retries before
escalation). The full extraction system prompt is reproduced in
Appendix~\ref{app:prompt_extraction}; the transcription and adjudication
prompts are in Appendices~\ref{app:prompt_transcription}
and~\ref{app:prompt_adjudication}.
Full model details are in Appendix~\ref{app:models}.
To maximize quality, we run extraction across three independent passes
and adjudicate disagreements
(adjudication verdict breakdown in Figure~\ref{fig:fields_and_adj}, right):

\emph{Pass~1 (PDF-direct):} Sonnet~4.6 reads each PDF directly via the
Anthropic Files API~\citep{anthropic2024filesapi}.

\emph{Pass~2 (text-based):} Sonnet~4.6 reads the Claude-transcribed
Markdown.

\emph{Pass~3 (comparison-prompt adjudication):} For the 3,401 records
(27.5\%) where the two passes disagreed on at least one key field, we
provide Opus~4.7 with the source text, both conflicting extractions, and
the specific disagreeing fields, and ask it to reason over the conflict
and produce the correct value.
This is methodologically stronger than majority voting: it is a
categorically different task (adjudication vs.\ extraction), the model
can identify cases where both passes were wrong, and it produces an
interpretable audit trail.
In 10.1\% of adjudicated records, Opus produced a value that neither
prior pass had reached.

\begin{figure}[t]
  \centering\includegraphics[width=\linewidth]{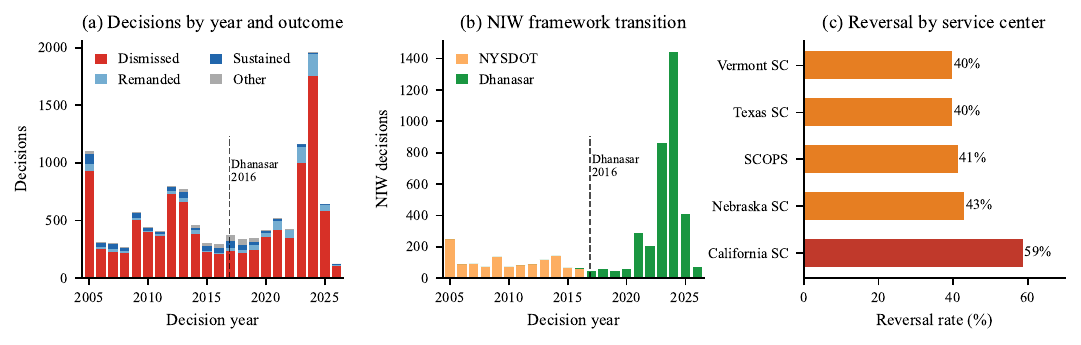}
  \caption{Corpus overview: (a) decisions per year colored by outcome;
    (b) NIW framework transition at December 2016 (\emph{Dhanasar});
    (c) reversal rate by originating USCIS Service Center.}
  \label{fig:corpus_overview}
\end{figure}

%% ====================================================================
\section{Dataset Analysis}
\label{sec:analysis}
%% ====================================================================

\begin{figure}[t]
  \centering\includegraphics[width=\linewidth]{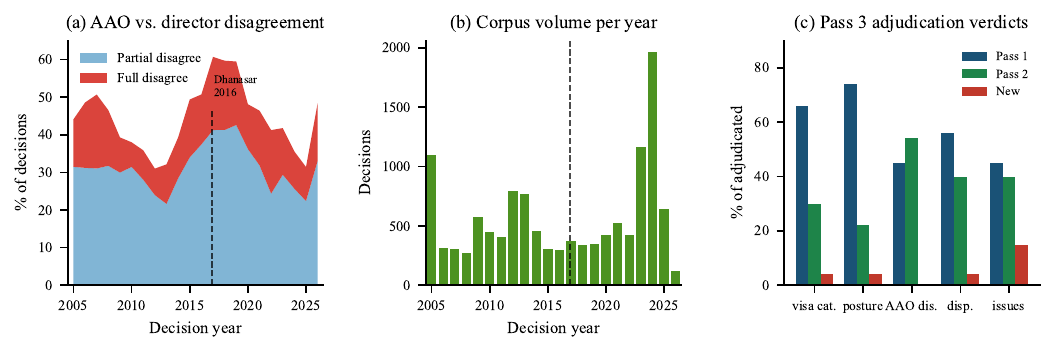}
  \caption{(a) AAO disagreement rate with the originating director over
    time. (b) Corpus volume per year. (c) Pass~3 adjudication verdicts:
    Opus~4.7 sided with each prior pipeline at roughly equal rates and
    produced new answers on 10.1\% of records.}
  \label{fig:temporal}
\end{figure}

\begin{figure}[t]
  \centering\includegraphics[width=\linewidth]{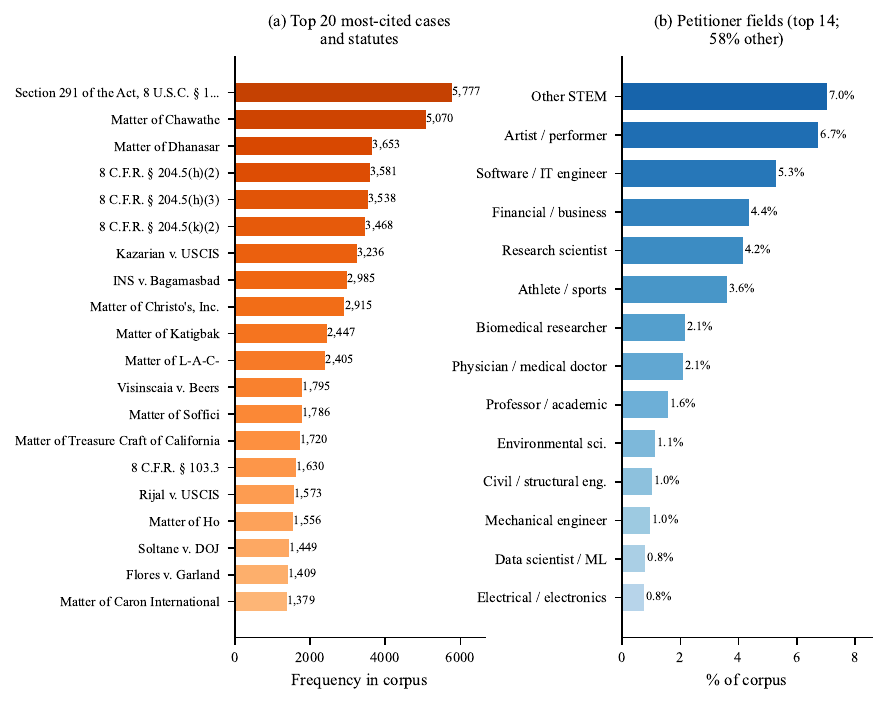}
  \caption{(a) Top 20 most-cited cases and statutes.
    \emph{Matter of Dhanasar} and \emph{Matter of Chawathe} dominate;
    the citation distribution provides a vocabulary for citation-grounding
    and hallucination studies.
    (b) Petitioner field distribution (top 14 categories; $\sim$15\%
    fall into a heterogeneous ``other'' group not shown).}
  \label{fig:fields_and_adj}
\end{figure}

\begin{figure}[t]
  \centering\includegraphics[width=\linewidth]{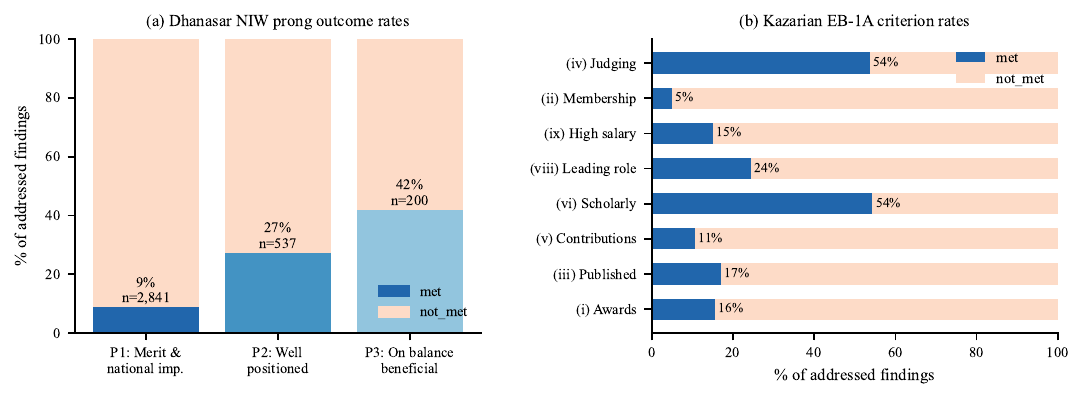}
  \caption{Per-prong and per-criterion outcome rates.
    (a) Dhanasar NIW prongs: Prong~3 (on balance beneficial) has the
    lowest petitioner win rate, consistent with its role as the most
    commonly dispositive prong in NIW reversals.
    (b) Kazarian EB-1A criteria: awards (i), published material (iii),
    and contributions (v) are the most contested.}
  \label{fig:prong_outcomes}
\end{figure}

\begin{figure}[t]
  \centering\includegraphics[width=\linewidth]{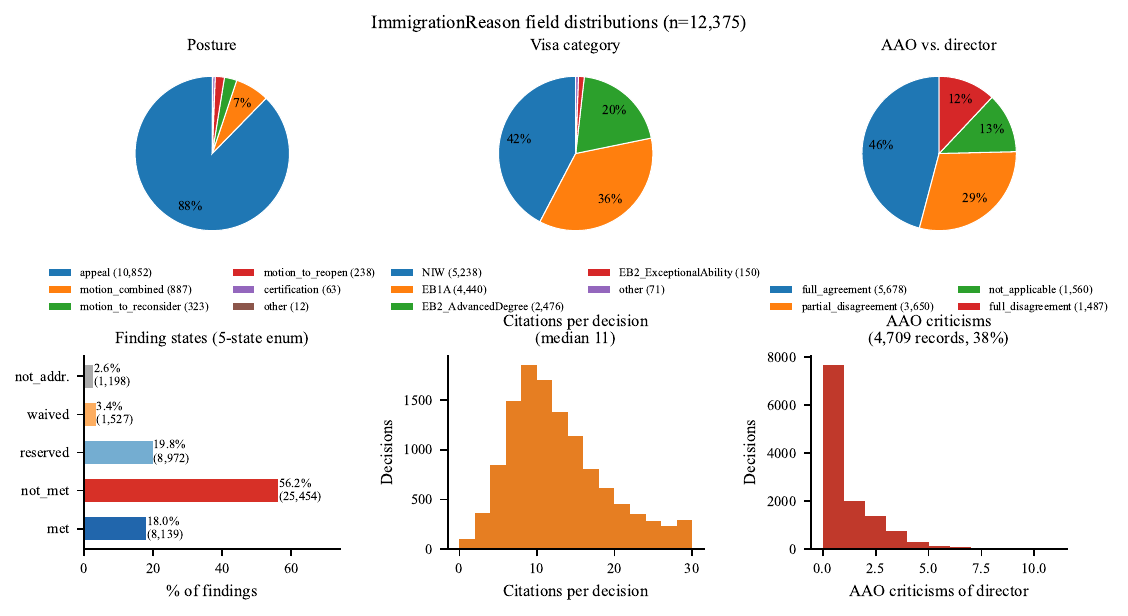}
  \caption{ImmigrationReason field distributions. Across posture,
    visa category, AAO stance, finding states, citations per decision,
    and AAO-criticisms-of-director counts.}
  \label{fig:schema_stats}
\end{figure}

Figures~\ref{fig:temporal} through \ref{fig:schema_stats} provide
a comprehensive statistical characterization of the dataset.

\paragraph{Form code is a noisy visa-category label.}
Of the 7,699 NIW-form-coded filings, only 5,238 (68\%) involve a
substantive NIW analysis; the remaining 32\% involve EB-2 qualification
thresholds, labor-certification disputes, ability-to-pay issues, or
procedural matters only.
Downstream users should use the LLM-extracted \texttt{visa\_category}
and \texttt{issue\_type} fields rather than the raw form code.

\paragraph{Reversal patterns by service center.}
Figure~\ref{fig:corpus_overview}(c) shows substantial variation in
reversal rates by originating service center.
California SC (54.4\%) and SCOPS (44.3\%) are reversed at notably higher
rates than Vermont SC (36.0\%).
This institutional-variation finding is only accessible via a structured
dataset; it cannot be derived from raw decision text alone.

\paragraph{Finding-state patterns.}
Figure~\ref{fig:prong_outcomes} shows that AAO finds petitioners succeed
on different criteria at very different rates.
For Kazarian EB-1A step-1, criteria (i) awards and (v) original
contributions have the lowest success rates; (iv) judging has among the
highest.
For Dhanasar NIW, Prong~3 (on balance beneficial) has the lowest success
rate among petitioners and is the most commonly reserved prong in
dismissals.

\paragraph{The natural regime-change experiment.}
The December 2016 \emph{Dhanasar} decision replaced the NYSDOT framework
for all NIW adjudications.
This transition appears as a sharp discontinuity in Figure~\ref{fig:corpus_overview}(b):
the corpus captures 1,280 NYSDOT-framework decisions (all before
December 2016) and 3,484 Dhanasar-framework decisions (all after), with
essentially zero overlap.
This creates a natural out-of-distribution split for any temporal
reasoning or legal-evolution study.

%% ====================================================================
\section{Quality Validation}
\label{sec:quality}
%% ====================================================================

\subsection{Three-Pass Extraction Agreement}

Table~\ref{tab:agreement} reports field-level agreement among the three
extraction passes over 12,372 paired records.

\begin{table}[t]
  \caption{Three-way field-level agreement. The improvement from legacy
    OCR (v1) to vision OCR (v2) against the PDF-direct reference
    validates that higher source-text quality improves extraction.}
  \label{tab:agreement}
  \centering\small
  \begin{tabular}{lcccccc}
    \toprule
    \textbf{Comparison} & \textbf{n} & \textbf{Visa cat.}
      & \textbf{Posture} & \textbf{AAO dis.} & \textbf{Dispositions}
      & \textbf{All 5} \\
    \midrule
    v1 (legacy OCR) vs.\ PDF & 12,372 & 98.2\% & 99.7\% & 89.5\% & 96.6\% & 71.0\% \\
    v2 (vision OCR) vs.\ PDF & 12,372 & 97.6\% & 99.8\% & 91.3\% & 97.2\% & 72.5\% \\
    v1 vs.\ v2 & 12,374 & 97.7\% & 99.7\% & 90.5\% & 97.0\% & 73.2\% \\
    \bottomrule
  \end{tabular}
\end{table}

Agreement on legally critical fields (visa category, posture,
dispositions) exceeds 97\%, confirming that independently-derived
extraction pipelines using different input modalities reach consistent
conclusions on the facts that matter most.
The lower agreement on \texttt{issue\_types} (83--85\%) reflects genuine
ambiguity in identifying which legal issues are substantively addressed,
exactly the cases resolved by Pass~3 adjudication.

\paragraph{Human expert verification.}
As an additional quality check, a stratified sample of 500 decisions
was reviewed by domain experts familiar with U.S.\ employment-based
immigration law.
The sample is stratified across visa category, era (pre/post-2017),
outcome, and extraction confidence.
Experts verified each extraction against the source text.
All reviewed samples were found to be correct and reasonable,
consistent with the quantitative agreement rates reported in
Table~\ref{tab:agreement}.

%% ====================================================================
\section{Dataset Contents: A Detailed Description}
\label{sec:contents}
%% ====================================================================

\paragraph{Source text (\texttt{aao\_text/clean/}).}
Clean Markdown files for all 12,375 decisions, organized by category and
filename stem.
Pre-2017 documents were transcribed via Claude~Sonnet~4.6 vision;
post-2017 text-native documents use PyMuPDF with
vision-transcription overrides for documents identified as
poor-quality.
Thirteen decisions (0.1\%) use legacy OCR fallback due to
content-filter interactions; these are flagged per-record.

\paragraph{Structured records (\texttt{corpus\_final.jsonl}).}
One JSON object per decision with the fields described in
Section~\ref{sec:schema}.
Key collection-level properties:
12,375 records; 45,290 per-criterion findings; $\sim$9,000 verbatim
director-criticism quotes; 150,000+ citations.
Each record includes a \texttt{\_provenance} field recording the
extraction model, prompt version, token counts, and which pipeline pass
produced the final record.

\paragraph{Extraction pipeline (\texttt{src/aao\_dataset/}).}
The full Python codebase for OCR, structured extraction, three-pass
validation, and quality reporting.
Includes the Pydantic schema, system prompts, batch-submission scripts,
and analysis notebooks.

\paragraph{OCR comparison (\texttt{ocr\_comparison.jsonl}).}
Per-record comparison between legacy OCR and Claude-transcribed text across
structural features, enabling reproducibility and future comparison
against alternative transcription methods.

\paragraph{Adjudication records (\texttt{adjudicated.jsonl}).}
For the 3,401 disagreement cases, the Opus~4.7 adjudication results
including the disagreeing fields and the adjudicator's final values,
enabling downstream analysis of where extraction pipelines tend to
disagree.

%% ====================================================================
\section{Proposed Research Directions}
\label{sec:directions}
%% ====================================================================

\ir{} is designed as a general-purpose resource. We describe several
research directions it enables.

\paragraph{\textbf{Outcome and Decision Prediction.}}

The dataset's outcome labels (sustained, remanded, dismissed) and
structured case facts enable outcome-prediction tasks at multiple
granularities: predicting the overall outcome given case facts,
predicting per-criterion findings given evidence descriptions, or
predicting which prong is most likely to be dispositive.
These tasks range in difficulty depending on how much of the decision
text is revealed to the model.
An important design consideration: modern AAO decisions state their
outcome in the introductory paragraph before the legal analysis;
a well-designed evaluation should strip this announcement to avoid
trivial text-extraction rather than genuine legal reasoning.

\paragraph{\textbf{Adjudicator Error Analysis and Detection.}}

The $\sim$9,000 verbatim AAO criticisms of originating-officer
decisions constitute a unique labeled corpus of adjudicator errors.
These range from misapplication of the legal standard
(``director imposed a novel evidentiary requirement contrary to
\emph{Kazarian}'')
to failure to consider submitted evidence
(``director did not address the petitioner's citation record'').
Researchers can:
(i)~cluster the criticisms into an empirical error typology using
embedding and clustering methods,
(ii)~train or evaluate models to detect officer-level errors given
denial reasoning, or
(iii)~study which types of errors are most prevalent by service center,
era, or petition category.
Given that USCIS is deploying AI in adjudication, error-detection
capabilities are directly safety-relevant: a model that can identify
the errors AAO catches in human officers can also flag analogous errors
in AI-generated denials.

\paragraph{\textbf{Legal Argument Quality and Reasoning Extraction.}}

The per-criterion structure of \ir{} enables fine-grained reasoning
extraction tasks: given a petitioner's evidence description and the legal
standard, predict whether AAO will find it sufficient.
More ambitiously, given the LAW section of a decision, generate the ANALYSIS
section and evaluate faithfulness by comparing extracted citations and
finding conclusions to the ground-truth record.
The distinction between \texttt{aao\_reasoning\_summary} (paraphrased)
and \texttt{aao\_reasoning\_quotes} (verbatim) enables both abstractive
and extractive approaches.

\paragraph{\textbf{Temporal Legal Dynamics.}}

The December 2016 \emph{Dhanasar} rule change is encoded in the data.
Researchers can study how models track legal-regime changes:
do models trained on pre-2016 data misapply the NYSDOT framework to
post-2016 cases?
Do models know which service center applied which framework in which year?
The dataset also spans 21 years of adjudicative practice, enabling
longitudinal studies of how evidentiary standards have evolved in
practice, independent of the formal framework change.

\paragraph{\textbf{Training, Fine-tuning, and Agent Design.}}

\ir{} can serve as training data for legal-reasoning models, not just
as an evaluation resource.
The paired (evidence-description, per-criterion-finding) data provides
supervised signal for evidence-sufficiency classification.
The director-reasoning + AAO-correction pairs provide training signal
for legal error detection.
The full decision text with structured annotations enables supervised
pretraining or fine-tuning on legal-reasoning tasks.
The multi-turn structure (petitioner argues, director responds, petitioner
appeals, AAO adjudicates) also provides rich trajectories for agent-design
research in legal advisory and petition-drafting contexts.

\paragraph{\textbf{Bias, Fairness, and Institutional Variation.}}

The dataset documents substantial variation in reversal rates across
USCIS Service Centers (ranging from 36.0\% to 54.4\%).
Combined with the petitioner-field annotations, it enables studies of
whether certain professions, petition types, or national-origin proxies
are systematically disadvantaged in adjudication.
The redaction of PII limits individual-level analysis but does not
preclude aggregate statistical studies at the petition-type or
service-center level.

\paragraph{\textbf{Hallucination and Citation Grounding.}}

The corpus-extracted citation lists provide ground truth for studying
hallucination in legal generation tasks.
Given the frequency distribution of citations
(Figure~\ref{fig:fields_and_adj}(b)),
researchers can examine whether models correctly cite \emph{Dhanasar},
\emph{Chawathe}, and other commonly-invoked precedents, or hallucinate
plausible-sounding but incorrect citations.

%% ====================================================================
\section{Limitations}
\label{sec:limitations}
%% ====================================================================

The corpus covers only employment-based immigration adjudication (EB-1A
and EB2-NIW), excluding asylum, removal, family-based, and non-U.S.\ systems,
so models trained here may not generalize to those domains.
AAO decisions are appellate, representing the subset of denials
petitioners chose to contest, which may not reflect initial-adjudication
behavior.
All PII has been redacted, precluding individual-level
demographic analysis.
Finally, while extraction is validated through three independent passes and
500 dedicated expert reviews, LLM-produced records may contain errors,
particularly on ambiguous older documents; the per-record
\texttt{extraction\_confidence} and \texttt{extraction\_notes} fields
surface known issues.

%% ====================================================================
\section{Conclusion}
\label{sec:conclusion}
%% ====================================================================

We present ImmigrationReason, a large-scale structured dataset of 12,375
U.S.\ administrative immigration appeals spanning 2005 to 2026.
The dataset provides deep annotations that do not exist in any prior
legal NLP resource: per-criterion evidence-sufficiency findings under a
five-state enum, verbatim adjudicator-criticism quotes, and explicit
director-vs.-AAO tracking at the criterion level.
Source text is provided as high-quality Claude-powered transcription,
validated against legacy OCR.
Extraction quality is validated through a three-pass pipeline and
500-record expert review.
ImmigrationReason enables a wide range of research in legal reasoning,
error detection, argument quality assessment, temporal legal dynamics,
and agent design for high-stakes regulatory domains.
We release all data, source text, extraction code, and schema under a
permissive license.

\begin{ack}[Anonymous for submission]\end{ack}

\bibliographystyle{plainnat}
\bibliography{references}

%% ====================================================================
\appendix
%% ====================================================================

\section{Extraction System Prompt (v0.3.1)}
\label{app:prompt_extraction}

The extraction system prompt is passed as a cached system block to every
call of Claude~Sonnet~4.6. It is reproduced in full below (line-continuation
backslashes omitted for readability).

\begin{small}
\begin{verbatim}
You are extracting structured data from USCIS Administrative Appeals
Office (AAO) non-precedent decisions for an academic legal-AI evaluation
dataset. Your job is to read one decision and produce a single structured
record matching the provided schema. The decisions cover EB-1A
(extraordinary ability) and NIW (national interest waiver) appeals, plus
motions and various procedural issues.

SOURCE-TEXT EXPECTATIONS (assume these -- do NOT narrate them)

Redactions: Every AAO decision is redacted by USCIS for anonymity. Names,
institutions, employers, journals, and other PII appear as [REDACTED],
blank boxes, or partial fragments. Do not mention redactions in
extraction_notes.

OCR artifacts (pre-2017): S may render as tj, _5_, or $; garbled words
(receet for receipt, goats for goals); page footers may interrupt flow.
Extract substance and ignore noise.

extraction_notes: Leave as null by default. Only set if a specific
extracted field is unreliable (e.g., "cite in footnote 3 is best-effort
due to OCR") or if unusual structure warrants flagging. NEVER write
notes that describe generic OCR quality, redactions, or restate schema
fields that already capture the information.

LEGAL FRAMEWORKS

NIW pre-2016: NYSDOT framework. Three factors: (1) substantial intrinsic
merit, (2) national in scope, (3) serves national interest substantially
greater than a U.S. worker. issue_type = niw_nysdot_analysis.

NIW post-2016: Dhanasar framework. Three prongs: (1) substantial merit
and national importance, (2) well-positioned to advance the endeavor,
(3) on balance beneficial to the U.S. issue_type = niw_dhanasar_analysis.

EB-1A: Two-step Kazarian framework. Step 1 = meet 3 of 10 criteria at
8 C.F.R. 204.5(h)(3)(i)-(x), issue_type = eb1a_kazarian_step1. Step 2 =
final merits determination, issue_type = eb1a_kazarian_step2.

PER-PRONG FINDINGS: CRITICAL DISTINCTIONS

aao_finding must be one of exactly:
- met / not_met: AAO evaluated the prong and reached a conclusion.
- reserved: AAO declines to reach due to another dispositive issue
  (Bagamasbad). Look for "we need not reach", "we hereby reserve".
- waived_by_petitioner: not raised on appeal, deemed abandoned.
- not_addressed: neither party discussed it at all.

DO NOT collapse reserved/waived/not_addressed into not_met.

Track director_finding and aao_finding SEPARATELY per prong. Record
aao_specific_criticism_of_director verbatim whenever AAO faults the
director's reasoning, even when affirming the same outcome.

FAITHFULNESS RULES

Quotes must be verbatim. Do not infer facts not stated. Do not
editorialize. The decision text is your only source of truth.

Now extract the decision.
\end{verbatim}
\end{small}

\section{Transcription Prompt}
\label{app:prompt_transcription}

The transcription prompt is used when calling Claude~Sonnet~4.6 to convert
scanned PDFs to Markdown. It is reproduced in full below.

\begin{small}
\begin{verbatim}
You are transcribing a USCIS Administrative Appeals Office (AAO)
non-precedent decision PDF into clean Markdown text. Produce a complete,
faithful transcription of every page in the document.

Rules:
1. Preserve ALL substantive text in document order: caption, header,
   case identifier, date, intro paragraphs, LAW section, ANALYSIS
   section (all sub-sections), CONCLUSION, and ORDER block(s).
2. Use Markdown structure: # / ## / ### for headings; **bold** for
   ORDER/FURTHER ORDER; _italic_ for case names.
3. Preserve redactions exactly (as [REDACTED], blank boxes, or dashes).
   Do not invent text where redactions are present.
4. Preserve all citations verbatim including I&N Dec., F.3d, U.S.C., and
   C.F.R. references.
5. Preserve footnote markers ([1], [2], etc.) and group footnote text at
   the end under a "Footnotes" heading.
6. OMIT page-number headers and embedded case-file numbers that interrupt
   paragraph flow (e.g., "Page 4", "EAC 03 052 50892").
7. Do NOT summarize, paraphrase, or correct apparent typos.
8. Output ONLY the transcribed Markdown. No commentary or preamble.

Critical completeness check: transcribe EVERY page end-to-end. Do not
stop after the cover letter or front matter. The decision body, ANALYSIS
section, CONCLUSION, and ORDER block come AFTER the cover materials and
MUST be included. Continue until you have transcribed the final ORDER block.
\end{verbatim}
\end{small}

\section{Adjudication Prompt Template}
\label{app:prompt_adjudication}

When Passes~1 and~2 disagree on one or more fields, the following context
block is appended to the source text before calling Opus~4.7. The
placeholders \texttt{\{pass1\_values\}} and \texttt{\{pass2\_values\}} are
filled with the actual extracted values for each disagreeing field.

\begin{small}
\begin{verbatim}
---
ADJUDICATION CONTEXT

Two prior extraction pipelines produced conflicting results for the
following fields. Re-read the source decision above and determine the
correct value for each disagreeing field.

Disagreeing fields:
  <field_label>:
    Pipeline 1 (PDF-direct):  <pass1_value>
    Pipeline 2 (text-based):  <pass2_value>
[... repeated for each disagreeing field ...]

Produce the fully correct structured extraction. For each disagreeing
field, identify which pipeline was correct, or provide the correct value
if both were wrong. Populate ALL fields (not just the disagreeing ones)
from the source text.
\end{verbatim}
\end{small}

\section{Model Information and Hyperparameters}
\label{app:models}

\begin{table}[h]
  \caption{Models and hyperparameters used across all pipeline stages.}
  \label{tab:models}
  \centering\small
  \begin{tabular}{llllp{3.8cm}}
    \toprule
    \textbf{Stage} & \textbf{Model} & \textbf{Temp.}
      & \textbf{Max tok.} & \textbf{Notes} \\
    \midrule
    Transcription & claude-sonnet-4-6 & 1.0 & 16,000
      & Files API; Batches API; 12,375 docs \\
    Pass 1 (PDF) & claude-sonnet-4-6 & 1.0 & 16,000
      & Files API; Batches API \\
    Pass 2 (text) & claude-sonnet-4-6 & 1.0 & 16,000
      & Sync parallel; 50 workers \\
    Pass 3 (adjudication) & claude-opus-4-7 & 1.0 & 16,000
      & Sync parallel; 40 workers; 3,401 docs \\
    \bottomrule
  \end{tabular}
\end{table}

Temperature 1.0 (the Anthropic API default) was used throughout and set
explicitly for Pass~3 to document that the adjudication is not a
deterministic repeat of Pass~2: stochastic sampling ensures genuine
independence.
Prompt caching reduced effective input cost by $\sim$70\% on cached
system-prompt tokens.
Total extraction cost was approximately \$1,300 USD using the Anthropic
Batches API (50\% discount on input and output tokens).

\section{Schema Enum Definitions}
\label{app:schema}

\textbf{IssueType (17 values):} \texttt{niw\_dhanasar\_analysis},
\texttt{niw\_nysdot\_analysis}, \texttt{eb1a\_kazarian\_step1},
\texttt{eb1a\_kazarian\_step2}, \texttt{eb2\_advanced\_degree\_qualification},
\texttt{eb2\_exceptional\_ability}, \texttt{labor\_cert\_validity},
\texttt{successor\_in\_interest}, \texttt{ability\_to\_pay},
\texttt{bona\_fide\_job\_offer},
\texttt{beneficiary\_qualifications\_per\_labor\_cert},
\texttt{motion\_requirements}, \texttt{appeal\_jurisdiction},
\texttt{fraud\_or\_misrepresentation}, \texttt{abandonment},
\texttt{constitutional\_or\_other\_legal\_error}, \texttt{other}.

\textbf{FindingResult (5 values):} \texttt{met}, \texttt{not\_met},
\texttt{reserved}, \texttt{waived\_by\_petitioner}, \texttt{not\_addressed}.

\textbf{Posture (8 values):} \texttt{appeal}, \texttt{motion\_to\_reopen},
\texttt{motion\_to\_reconsider}, \texttt{motion\_combined},
\texttt{appeal\_of\_approval}, \texttt{certification},
\texttt{remand\_review}, \texttt{other}.

\textbf{Disposition (14 values):} \texttt{appeal\_sustained},
\texttt{appeal\_dismissed}, \texttt{appeal\_summarily\_dismissed},
\texttt{appeal\_dismissed\_as\_moot},
\texttt{appeal\_dismissed\_as\_abandoned},
\texttt{appeal\_rejected\_improperly\_filed},
\texttt{appeal\_withdrawn}, \texttt{motion\_granted},
\texttt{motion\_dismissed}, \texttt{motion\_denied},
\texttt{motion\_rejected}, \texttt{directors\_decision\_withdrawn},
\texttt{remanded}, \texttt{fraud\_finding}.

\section{Dataset Comparison}
\label{app:comparison}

Table~\ref{tab:comparison} compares ImmigrationReason with related
legal NLP resources. The key differentiator is structured per-criterion
annotations under a five-category label; no prior resource provides this.

\begin{table}[h]
  \caption{ImmigrationReason vs.\ related legal NLP datasets.}
  \label{tab:comparison}
  \centering\small
  \begin{tabular}{p{3.4cm}rp{1.4cm}p{1.4cm}p{1.6cm}}
    \toprule
    \textbf{Dataset} & \textbf{Size}
      & \textbf{Structured annot.}
      & \textbf{Per-criterion findings}
      & \textbf{US admin. law} \\
    \midrule
    \textbf{ImmigrationReason (ours)} & \textbf{12,375}
      & \textbf{yes} & \textbf{yes} & \textbf{yes} \\
    LegalBench~\citep{guha2023legalbench}      & 53,000  & no  & no  & no  \\
    LexGLUE~\citep{chalkidis2022lexglue}       & 31,000  & no  & no  & no  \\
    Pile of Law~\citep{hendersonkrass2022pile}  & 256M    & no  & no  & no  \\
    CaseHOLD~\citep{zheng2021does}             & 53,000  & no  & no  & no  \\
    LawBench~\citep{fei2023lawbench}           & 21,000  & no  & no  & no  \\
    Refugee NLP~\citep{barale2023refugee}      & 400     & no  & no  & no  \\
    \bottomrule
  \end{tabular}
\end{table}

\section{Data Card}
\label{app:datacard}

\paragraph{Dataset summary.}
\ir{} contains 12,375 non-precedent decisions of the USCIS Administrative
Appeals Office (AAO) for EB-1A extraordinary-ability and NIW national-interest-waiver
petitions, spanning 2005--2026.
Each record carries structured extraction (legal frameworks, per-criterion findings,
citations, final dispositions) plus the full decision text as a Claude-transcribed
Markdown document.

\paragraph{Intended uses.}
The dataset is intended for legal NLP research, including:
outcome prediction and adjudicator-error analysis on administrative law;
evaluation of LLM legal reasoning at the criterion level;
study of how legal-regime transitions (e.g., the 2016 \emph{Dhanasar} change)
affect adjudication patterns;
and development of AI agents for high-stakes regulatory domains.

\paragraph{Out-of-scope uses.}
\ir{} is not intended for use as legal advice, for providing guidance on
specific pending immigration petitions, or as the basis for any automated
adjudication system.
The dataset documents historical AAO decisions; it does not predict or prescribe
outcomes for future cases.

\paragraph{Dataset composition.}
The corpus comprises 12,375 non-precedent AAO decisions: 6,981 NIW and
5,394 EB-1A (by form-code category).
Each decision is represented by one row in \texttt{decisions.parquet},
with nested structs for legal issues, per-criterion findings, citations,
and final orders.
The flat \texttt{findings\_long.parquet} table contains 45,290 criterion-level
rows derived from those nested structures.
Source text is included directly in the \texttt{text} column.
No personally identifiable information is included.

\paragraph{Collection process.}
Source PDFs were collected from the public USCIS AAO website.
All documents are non-precedent decisions and are U.S.\ Government public records.
Structured extraction was performed via a three-pass LLM pipeline
(Section~\ref{sec:pipeline}); the full codebase is released alongside the data.

\paragraph{Preprocessing and transcription.}
Post-2017 documents were processed with PyMuPDF (embedded text layer).
Pre-2017 scanned documents were transcribed with Claude Sonnet~4.6 using
a detailed transcription prompt (Appendix~\ref{app:prompt_transcription}).
Transcription quality was validated against legacy OCR across 12,375 paired
records (Table~\ref{tab:ocr}).

\paragraph{Known limitations and biases.}
\textit{Appellate selection bias:} the corpus covers only cases where the initial
petition was denied by a USCIS service center \emph{and} the petitioner chose to
appeal to the AAO.
Approvals at the service center level are not represented.
\textit{Non-precedent decisions:} these decisions are not binding on future
adjudications; doctrinal authority rests with AAO precedent decisions and federal
court holdings not covered here.
\textit{Extraction errors:} while expert review found all sampled records correct
and reasonable, LLM-produced extractions may contain errors at low frequency,
particularly for unusual procedural postures or ambiguous rulings.

\paragraph{Social impact.}
The dataset enables researchers to study patterns in administrative immigration
adjudication at scale, including criterion-level error rates, the impact of legal
reforms, and geographic variation in service-center outcomes.
We note the dual-use risk: detailed analysis of adjudication patterns could
be used to craft petitions in ways that exploit identified weaknesses in the
adjudication process rather than genuinely satisfying the legal standard.
We release the dataset for research purposes and encourage responsible use.

\paragraph{Maintenance.}
The dataset reflects AAO non-precedent decisions available as of early 2026.
The authors intend to release annual updates as new decisions are published.
Issues and corrections can be reported via the HuggingFace dataset repository.

\paragraph{License.}
Structured annotations and extraction records are released under
CC-BY~4.0~(\url{https://creativecommons.org/licenses/by/4.0/}).
Source text (decision transcriptions) are U.S.\ Government public records
and are in the public domain.

\section{Croissant Metadata}
\label{app:croissant}

\ir{} is released with Croissant~1.0 JSON-LD metadata~\citep{mlcroissant2024},
available at the HuggingFace repository as \texttt{croissant.json}.
Below is an abbreviated excerpt showing the dataset-level metadata and
the \texttt{decisions} RecordSet; the full file covers all three
parquet tables and all fields.

\begin{verbatim}
{
  "@context": {
    "@vocab": "https://schema.org/",
    "cr": "http://mlcommons.org/croissant/"
  },
  "@type": "sc:Dataset",
  "cr:conformsTo": "http://mlcommons.org/croissant/1.0",
  "name": "ImmigrationReason",
  "url": "https://huggingface.co/datasets/afsharrad/immigration-reason",
  "license": "https://creativecommons.org/licenses/by/4.0/",
  "version": "1.0.0",
  "distribution": [
    { "@type": "cr:FileObject", "@id": "decisions-parquet",
      "name": "decisions.parquet",
      "encodingFormat": "application/x-parquet" },
    { "@type": "cr:FileObject", "@id": "findings-parquet",
      "name": "findings_long.parquet",
      "encodingFormat": "application/x-parquet" },
    { "@type": "cr:FileObject", "@id": "ocr-parquet",
      "name": "ocr_comparison.parquet",
      "encodingFormat": "application/x-parquet" }
  ],
  "recordSet": [
    { "@type": "cr:RecordSet", "@id": "decisions",
      "field": [
        { "name": "filename_stem", "dataType": "sc:Text" },
        { "name": "category",      "dataType": "sc:Text" },
        { "name": "decision_year", "dataType": "sc:Integer" },
        { "name": "posture",       "dataType": "sc:Text" },
        { "name": "legal_issues",  "dataType": "sc:Text" },
        { "name": "aao_finding",   "dataType": "sc:Text" },
        { "name": "text",          "dataType": "sc:Text" }
      ]
    }
  ]
}
\end{verbatim}

\end{document}